\documentclass[final]{cvpr}

\usepackage{times}
\usepackage{epsfig}
\usepackage{graphicx}
\usepackage{amsmath}
\usepackage{amssymb}

\usepackage{subcaption}
\usepackage{multirow}
\usepackage{makecell}
\usepackage{booktabs}
\usepackage{enumitem}
\usepackage{float}

\usepackage{paralist}

\usepackage[pagebackref=true,breaklinks=true,colorlinks,bookmarks=false]{hyperref}

\def\cvprPaperID{5291} 
\def\confYear{CVPR 2021}

\begin{document}
	


\title{Event-based Synthetic Aperture Imaging with a Hybrid Network}

\author{Xiang Zhang\footnotemark[1],$\ $  Wei Liao\footnotemark[1],$\ $  Lei Yu\footnotemark[2],$\ $  Wen Yang,$\ $ Gui-Song Xia\\
	Wuhan University, Wuhan, China.
	\\
	{\tt\small \{xiangz, weiliao, ly.wd, yangwen, guisong.xia\}@whu.edu.cn}
}

\maketitle
\begin{abstract}
	Synthetic aperture imaging (SAI) is able to achieve the \textbf{see through} effect by blurring out the off-focus foreground occlusions and reconstructing the in-focus occluded targets from multi-view images. However, very dense occlusions and extreme lighting conditions may bring significant disturbances to the SAI based on conventional frame-based cameras, leading to performance degeneration. To address these problems, we propose a novel SAI system based on the event camera which can produce asynchronous events with extremely low latency and high dynamic range. Thus, it can eliminate the interference of dense occlusions by measuring with almost continuous views, and simultaneously tackle the over/under exposure problems. To reconstruct the occluded targets, we propose a hybrid encoder-decoder network composed of spiking neural networks (SNNs) and convolutional neural networks (CNNs). In the hybrid network, the spatio-temporal information of the collected events is first encoded by SNN layers, and then transformed to the visual image of the occluded targets by a style-transfer CNN decoder. Through experiments, the proposed method shows remarkable performance in dealing with very dense occlusions and extreme lighting conditions, and high quality visual images can be reconstructed using pure event data.
	
	
	%
	%
	%
	%
\end{abstract}

\section{Introduction}\label{chapter-1}
\footnotetext[2]{Corresponding author}
\footnotetext[1]{Equal contribution}
\footnotetext{The research was partially supported by the National Natural Science Foundation of China under Grants 61871297, the National Natural Science Foundation of China Enterprise Innovation Development Key Project under Grant U19B2004 and the Fundamental Research Funds for the Central University under Grant 2042020kf0019.}
\begin{figure}[!t]
	\centering
	\includegraphics[width=0.97\linewidth]{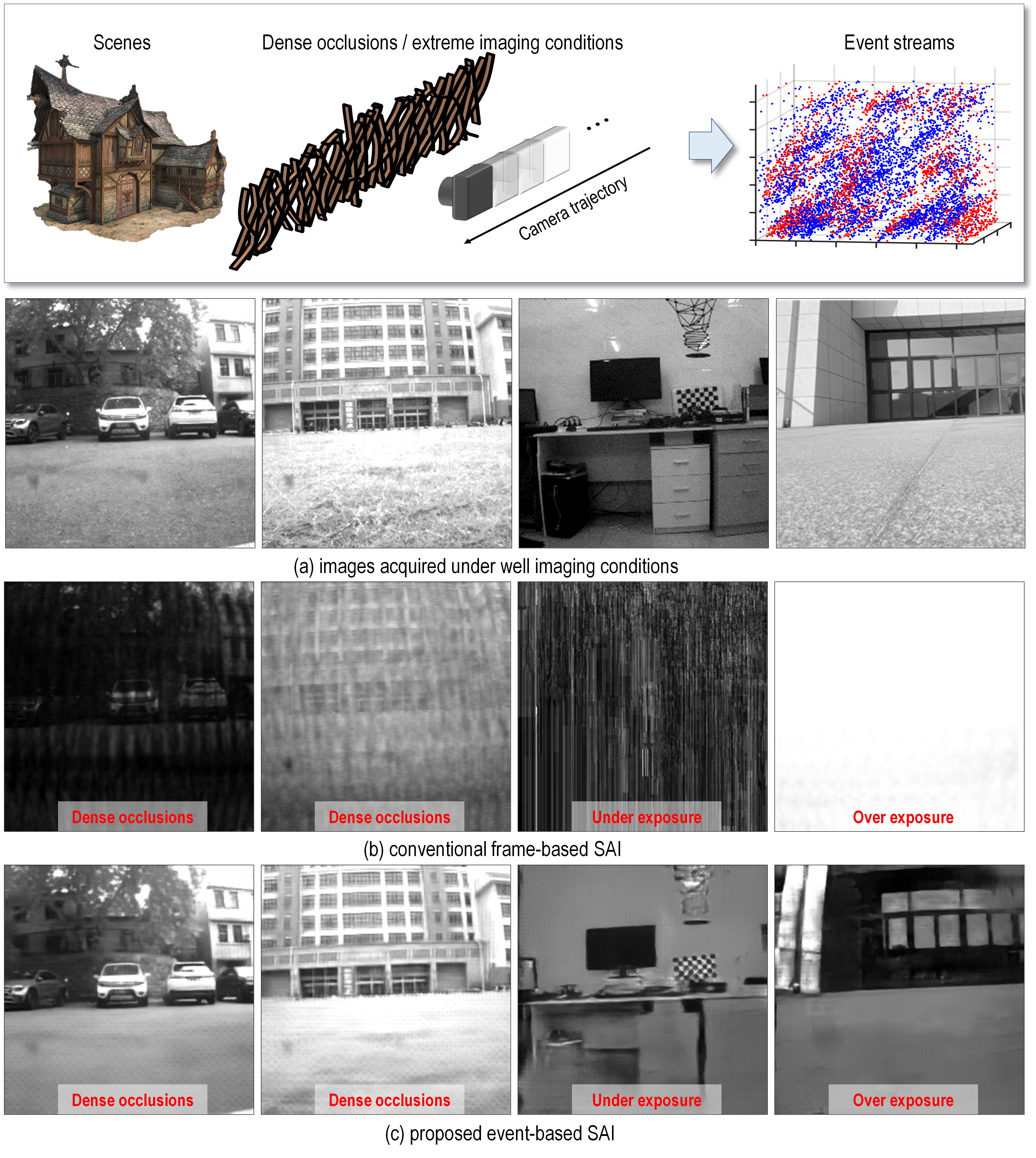}
	\vspace{-2mm}
	\caption{Prototype of the \textbf{event-based synthetic aperture imaging (E-SAI)} system and the illustrative examples of see through imaging under harsh environments via (b) the conventional frame-based SAI and (c) the event-based SAI. Under either very dense occlusions or extreme lighting scenes, the proposed E-SAI method can successfully generate high quality visual images for the occluded 
		scenes. }
	\label{First}
\end{figure}
Harsh environments, \eg with dense occlusions and extreme lighting conditions, often prohibit the efficient imaging of real scenes, due to the fact that the collected light information is very limited and moreover severely disturbed. Synthetic aperture imaging (SAI) tackles the problem of \emph{seeing through} occlusions via multi-view exposures \cite{vaishUsingPlaneParallax2004,vaishReconstructingOccludedSurfaces2006}, forming the light field \cite{ZHANG2017175} of the target scene under occlusions. The basic idea of SAI is to extract the light information of the occluded scenes while filter out foreground occlusions \cite{peiSyntheticApertureImaging2013,xiaoSeeingForegroundOcclusion2017}.
However, very dense occlusions and extreme lighting scenes may bring severe disturbances, leading to serious degradation on imaging quality or even failure reconstructions (\eg, Fig.~\ref{First}). 
\begin{itemize}
	\item \textbf{Very dense occlusions:}  With conventional frame-based cameras, the light cues are captured via brightness intensities. Very dense occlusions will greatly decrease the ``signal'', \ie the light from target scenes, while increase the ``noise'', \ie disturbances from foreground occlusions, leading to considerable reduction of the Light-SNR (ratio of ``signal'' to ``noise'').
	\item \textbf{Extreme lighting scenes:} Due to the low dynamic range (\eg about $60$ dB), images from conventional frame-based cameras usually suffer from the over/under exposure problems under extreme lighting conditions. It will severely degrade the imaging quality and thus reduce the confidence of the light information from target scenes. 
\end{itemize}
As a consequence, conventional frame-based SAI (F-SAI) often fails in these cases, and it is of great demand to develop new SAI methods to handle such harsh environments.

\par 
\par 
In this paper, we address the aforementioned problems by presenting a novel SAI method with event cameras~\cite{9138762}. Event cameras only measure the pixel-wise brightness changes of scenes in an asynchronous manner, leading to many outstanding properties including extremely low latency (in the order of $\mu$s), high dynamic range ($>$ 120 dB) and low power consumption \cite{lichtsteiner128Times1282008,9138762}.
Instead of using frame-based intensity images, as shown in Fig.~\ref{First}, event-based SAI (E-SAI) collects the light information from occluded targets via event streams, representing the brightness difference between the foreground occlusions and the occluded targets. This mechanism means that a higher density of occasions produces more events from occluded targets, \ie more light information of targets can be recorded. With the low latency, event cameras can capture adequate information of the occluded object from almost continuous viewpoints. Thanks to the high dynamic range of event camera, E-SAI is able to collect confident light information from occluded targets even under extreme lighting conditions, making the reconstruction of scenes feasible (\eg, Fig.~\ref{First}). 
\par 
Although E-SAI can easily handle the aforementioned problems, we still have to answer the following question: \textit{how to effectively process the event stream and reconstruct the high quality visual images of occluded targets?} Since the working mechanism of event camera differs radically from that of the frame-based one, conventional computer vision methods, \eg convolutional neural networks (CNNs), cannot be directly applied to such asynchronous event streams, where the temporal and spatial information of events should be simultaneously considered \cite{9138762}.
\par 
The spiking neural network (SNN) \cite{maass1998pulsed,maassNetworksSpikingNeurons1997} serves as a perfect model for integrating spatio-temporal information. Unlike other artificial neural networks, spiking neurons
do not respond to stimulus in a synchronous fashion. Instead, the membrane potential of spiking neurons updates over time, and a spike will be generated whenever the membrane potential exceeds a specific spiking threshold. Thus the spatio-temporal information is naturally encoded in the spike position and timing. Exploiting this, the influence of noise events can be further mitigated from the temporal dimension, leading to the improvement of Light-SNR. However, recent researches have observed the vanishing spike phenomenon \cite{pandaScalableEfficientAccurate2020} in deep spiking layers. Thus SNNs often suffer from performance degradation when the number of layers increases.
\par 
To tackle this, we propose a hybrid neural network that contains a SNN encoder and a CNN decoder. 
With initial spiking layers, the spatio-temporal information of events can be efficiently integrated and encoded. 
Then, the CNN is able to decode the rich output of SNN, and effectively reconstruct the visual image of occluded targets.
Therefore, this architecture not only utilizes sufficient information of events, but also guarantees the overall performance of reconstruction. 
\par 
In a nutshell, contributions of this paper are three-fold:
\begin{itemize}
	\item We present a novel event-based SAI algorithm with systematic analysis, which can overcome the dilemma that the conventional F-SAI faces under very dense occlusions and extreme lighting conditions.
	\item We propose a hybrid SNN-CNN encoder-decoder network to reconstruct high quality visual images for E-SAI. By leveraging the merits of SNN and CNN, the spatio-temporal information of events can be well retained and utilized, and thus the occluded target can be effectively reconstructed.
	\item We construct an event-based SAI dataset to evaluate the proposed method, and make them available to the research community.
\end{itemize}

\begin{figure*}[!t]
	\centering
	\includegraphics[width=0.97\linewidth]{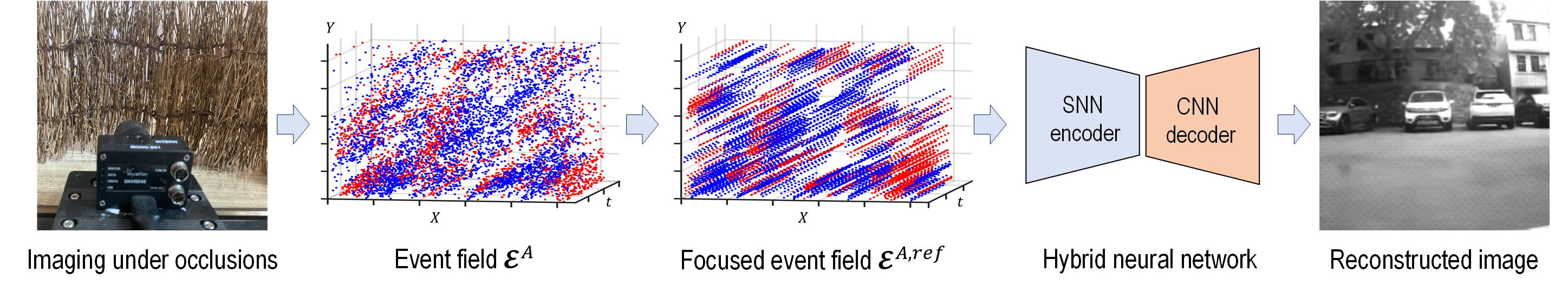}
	\vspace{-2mm}
	\caption{Overall pipeline of the proposed E-SAI. As moving the event camera, E-SAI collects event streams $\mathcal{E}_\theta^A$ with almost continuous viewpoints $\theta$ and forms the \emph{event field} $\boldsymbol{\mathcal{E}}^A$. To reconstruct high quality images from $\boldsymbol{\mathcal{E}}^A$, we propose to employ the hybrid SNN-CNN network after the event refocusing process. }
	\label{pipline}
	\vspace{-2mm}
\end{figure*}

\section{Related Work}\label{chapter-2}
\textbf{Synthetic Aperture Imaging:} How to see through the foreground occlusion has attracted considerable interest for decades \cite{vaishUsingPlaneParallax2004,yangContinuouslyTrackingSeethrough2011,peiSyntheticApertureImaging2013,xiaoSeeingForegroundOcclusion2017}.
Through calibrating the images captured by camera arrays, a plane + parallax framework was proposed to solve the de-occlusion problem \cite{vaishUsingPlaneParallax2004}.
Since the output of camera arrays can be regarded as a virtual camera imaging with large-aperture lens, the foreground occlusion can be effectively blurred out when the background target is refocused on. But this method often results in blurry images because the information from both occlusions and targets will be indiscriminately used for reconstruction. 
To improve the performance, a variety of techniques have been exploited to filter the disturbance of occlusions, including depth-based approach \cite{yangContinuouslyTrackingSeethrough2011}, energy minimization \cite{peiSyntheticApertureImaging2013} and k-means clustering \cite{xiaoSeeingForegroundOcclusion2017}. By separating targets from foreground occlusions, a better de-occlusion effect can be achieved using only target information. 
\par 
The principle of traditional F-SAI is to reconstruct the occluded target via multi-view images captured by a moving camera \cite{ZHANG2017175} or a camera array system \cite{10.1145/1186822.1073259,wangDeOccNetLearningSee2019}.
By projecting all images to the plane where targets are located, the light information of occluded target is aligned while the occlusion becomes out of focus. Afterward, reconstruction can be performed to achieve the see through effect.
But due to the inherent mechanism of traditional camera, the Light-SNR of captured images is often severely reduced when encountering very dense occlusions or extreme light scenes, resulting in significant performance degradation.

\textbf{Event Cameras:} Instead of frame-based intensity images, event cameras generate asynchronous events \cite{9138762}, composed of pixel position, time stamp and polarity. Specifically, the $i$-th event $e_i = (p_i,x_i,t_i)$ is triggered at pixel position $x_i$ and time $t_i$ whenever the log-scale brightness change exceeds a pre-setting threshold $\eta$, \ie
\begin{equation}\label{logchange}
\operatorname{log}(I(x_i,t_i)) - \operatorname{log}(I(x_i,t_i-\Delta t_i)) \geq p_i\cdot \eta,
\end{equation}
where $I(\cdot)$ denotes the intensity of pixel; $\Delta t_i$ indicates the time since the last event at position $x_i$; $p_i\in\{+1,-1\}$ is the polarity representing the sign of brightness change \cite{lichtsteiner128Times1282008}. This paradigm shift in visual information acquisition leads to many outstanding properties like extremely low latency and high dynamic range, and promotes great potential in many computer vision tasks like optical flow estimation \cite{wang2020event}, high dynamic range (HDR) imaging \cite{rebecqEventsToVideoBringingModern2019} and simultaneous localization and mapping (SLAM) \cite{vidalUltimateSLAMCombining2018}.


Similarly, event cameras pose great advantages in dealing with the see through tasks.
Due to the low latency property, sufficient light information of occluded targets can be acquired by event cameras under disturbance of dense occlusions.
On the other hand, the high dynamic range of event cameras provides the possibility of measuring light information under extreme lighting conditions. 
Thus it motivates us to exploit event cameras to tackle the problem of SAI under very dense occlusions and extreme lighting conditions, and propose the E-SAI.

\section{Problem Statement}
Suppose for some static unknown scene $A$ with $I_\theta^A(u,v)$ representing the projected brightness intensity captured with the camera pose $\theta$,  where $u,v$ are respectively horizontal / vertical coordinates. Then $\boldsymbol{I}^A\triangleq \{I_\theta^A\}_{\theta\in \mathcal{P}}$ forms a tensor of  light field of $A$ with $\mathcal{P}$ the set of camera poses. Analogically, the light field of occlusions $O$ can be represented as  $\boldsymbol{I}^O\triangleq \{I_\theta^O\}_{\theta\in \mathcal{P}}$ with $I_\theta^O(u,v)$ denoting the brightness intensity captured with the camera pose $\theta$. 

\textbf{F-SAI:} The task of F-SAI is to achieve the see through imaging from limited number of occluded observations, \ie, $\boldsymbol{\bar I}^A =  \{\bar I^A_\theta\}_{\theta\in \mathcal{P}}$ with $|\mathcal{P}|<\infty$ and 
\begin{equation}\label{FSAI}
\bar I^A_\theta = \mathcal{M}^O(I_\theta^A) + I_\theta^O + I^n
\end{equation}
where $I^n$ denotes the measurement noises and $\mathcal{M}^O$ represents the masking operator for $\mathcal{M}^O(\cdot)=0$ only when it is occluded by $O$. With very dense occlusions, $\bar I^A_\theta$ will be severely contaminated and the extreme lighting conditions make the observations completely saturated and incorrect, leading to failure reconstruction of visual images for $A$.
\par 
\textbf{E-SAI:} As illustrated in Fig.~\ref{First}, events are induced by the brightness change as moving the event camera, then we can denote the collected events with camera pose $\theta$ as a set of stream $\mathcal{E}_\theta^A\triangleq \{e_i\}_{i=1}^{M}=\{(p_i,x_i,t_i)\}_{i=1}^{M}$ with $M=|\mathcal{E}_\theta^A|$. According to the generating process, we can divide $\mathcal{E}_\theta^A$ into two categories:
(1) \emph{Signal events}, denoted as $\mathcal{E}^{OA}_\theta$, are induced by the brightness difference between the scene $A$ and the occlusion $O$. Then based on Eq.~\eqref{logchange}, the number of events emitted for $\mathcal{E}^{OA}_\theta$,
\begin{equation}\label{logdiff}
|\mathcal{E}^{OA}_\theta|  \propto  \left\vert \log (I^A_\theta) - \log (I^O_\theta)  \right\vert.
\end{equation}
(2) \emph{Noise events} include the physical noises $\mathcal{E}^{n}$ inherently from the event camera and the interference events induced by the brightness change (caused by textures) of occlusions $\mathcal{E}^{OO}_\theta$ and occluded targets $\mathcal{E}^{AA}_\theta$ as moving the event camera.


Due to the low latency property, E-SAI is able to collect events $\mathcal{E}^A_\theta$ from almost continuous viewpoints $\theta$ and form the \emph{event field}, \ie  $\boldsymbol{\mathcal{E}}^A =  \{\mathcal{E}^A_\theta\}_{\theta\in \mathcal{P}}$, with $|\mathcal{P}|\to\infty$ and
\begin{equation}\label{ESAI}
\mathcal{E}^A_\theta  = \mathcal{E}^{OA}_\theta + {\mathcal{E}^{OO}_\theta + \mathcal{E}^{AA}_\theta +\mathcal{E}^{n}}
\end{equation}
where $\mathcal{E}^{OA}_\theta$ encodes the light information from the occluded targets and the other terms are considered as noises. Thus the main problem of E-SAI is to reconstruct the high quality visual images of the scene $A$ from the event field $\boldsymbol{\mathcal{E}}^A$ which are severely disturbed by noise events.

\section{Event-based SAI}\label{sec3-1}
Fig.~\ref{pipline} illustrates the overall pipeline of the proposed E-SAI algorithm, which aims at reconstructing the high quality visual images of occluded targets. It consists of two main steps: \emph{refocusing} and \emph{reconstruction}. 
The purpose of refocusing is to align the signal events, while scatter out the noise events from both spatial and temporal dimensions. For the reconstruction, a hybrid SNN-CNN network is proposed to mitigate the disturbance of noise mentioned in Eq.~(\ref{ESAI}). With spiking layers, the influence of scattered noise events can be eliminated from the temporal dimension, thus a clean visual result can be decoded by the CNN.
\begin{figure}[t]
	\centering
	\includegraphics[width=0.87\linewidth]{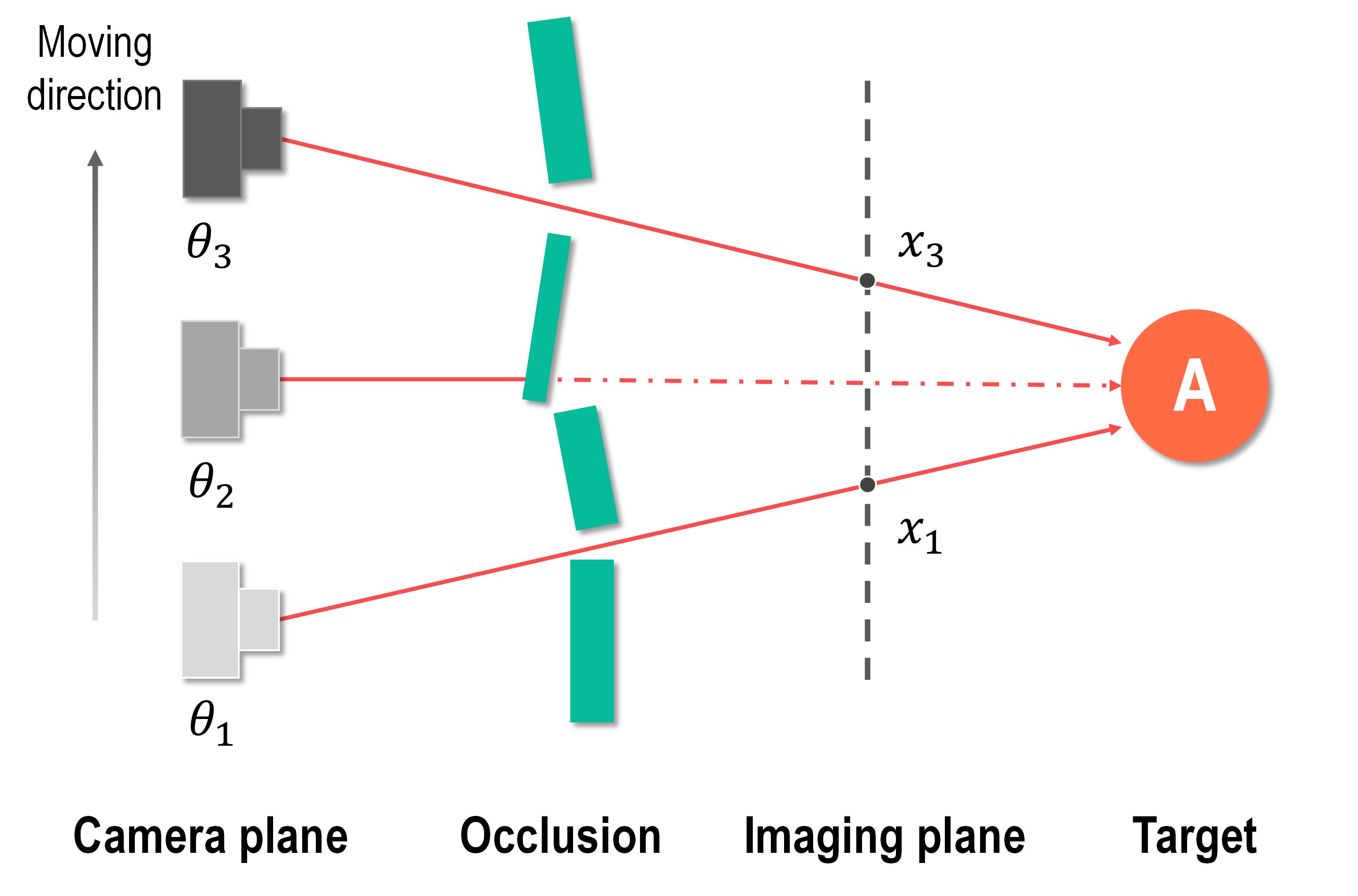}
	\vspace{-2mm}
	\caption{Diagram of event generating process in E-SAI system. As the event camera moves, signal events triggered with camera poses $\theta_1$ and $\theta_3$ are induced by the brightness difference between occlusions (\textcolor{green}{green}) and the target $A$ (\textcolor[RGB]{255,107,66}{orange}), while noise events triggered with camera pose $\theta_2$ are induced by textures on occlusions. }
	\label{refocus}
	\vspace{-3mm}
\end{figure}
\subsection{Event Refocusing}
Previous works \cite{Gallego_2018_CVPR,rebecq2018emvs} have presented the similar ideas of event refocusing. But in our case, the basic principles of the aforementioned techniques are violated due to the disturbance of dense occlusions and extreme lighting scenes. Thus we only consider a simple situation with linear camera motion and known target depth in this work. As displayed in Fig.~\ref{refocus}, a moving camera is employed to collect events $\boldsymbol{\mathcal{E}}^A$ from multiple viewpoints. We assume that the event camera keeps staying on the camera plane and the optical axes of all camera poses are parallel. 
Since all the events are emitted asynchronously as the camera moves, a pixel-wise refocusing process needs to be performed for event alignment.
Define $\theta^{ref}$ as the reference camera pose and $X_{i}$ as the coordinate of pixel $x_i$ in the camera coordinate system at pose $\theta_{i}$. According to the multiple view geometry \cite{hartley2003multiple} and the pinhole imaging model \cite{PEI20121637}, the refocusing equation can be formulated as \cite{AAS-CN-2020-0388}:
\begin{equation}\label{trans-final}
x_i^{ref} = KR_{i}K^{-1}x_{i}+\frac{KT_{i}}{d},
\end{equation}
where $x_i^{ref}$ represents the target pixel position on the reference imaging plane; $K$ is the intrinsic matrix of camera; $R_{i}, T_{i}$ are the rotation and translation matrices between camera poses $\theta_i$ and $\theta^{ref}$; target depth $d$ is the distance between target $A$ and the camera plane.
\par 
Exploiting Eq.~(\ref{trans-final}), the refocused event field can be obtained $\boldsymbol{\mathcal{E}}^{A,ref} =  \{\mathcal{E}^{A,ref}_\theta\}_{\theta\in \mathcal{P}} $ where $\mathcal{E}^{A,ref}_\theta = \{e_i^{ref}\}_{i=1}^{M}=\{(p_i,x^{ref}_i,t_i)\}_{i=1}^{M}$ and all events are projected to the imaging plane of the reference camera at pose $\theta^{ref}$. After refocusing, the events triggered by target $A$ are successfully aligned, while others, \eg the events generated by occlusions, are scattered out in both temporal and spatial dimensions, achieving a preliminary de-occlusion effect.


\begin{figure}[t]
	\centering
	\includegraphics[width=0.92\linewidth]{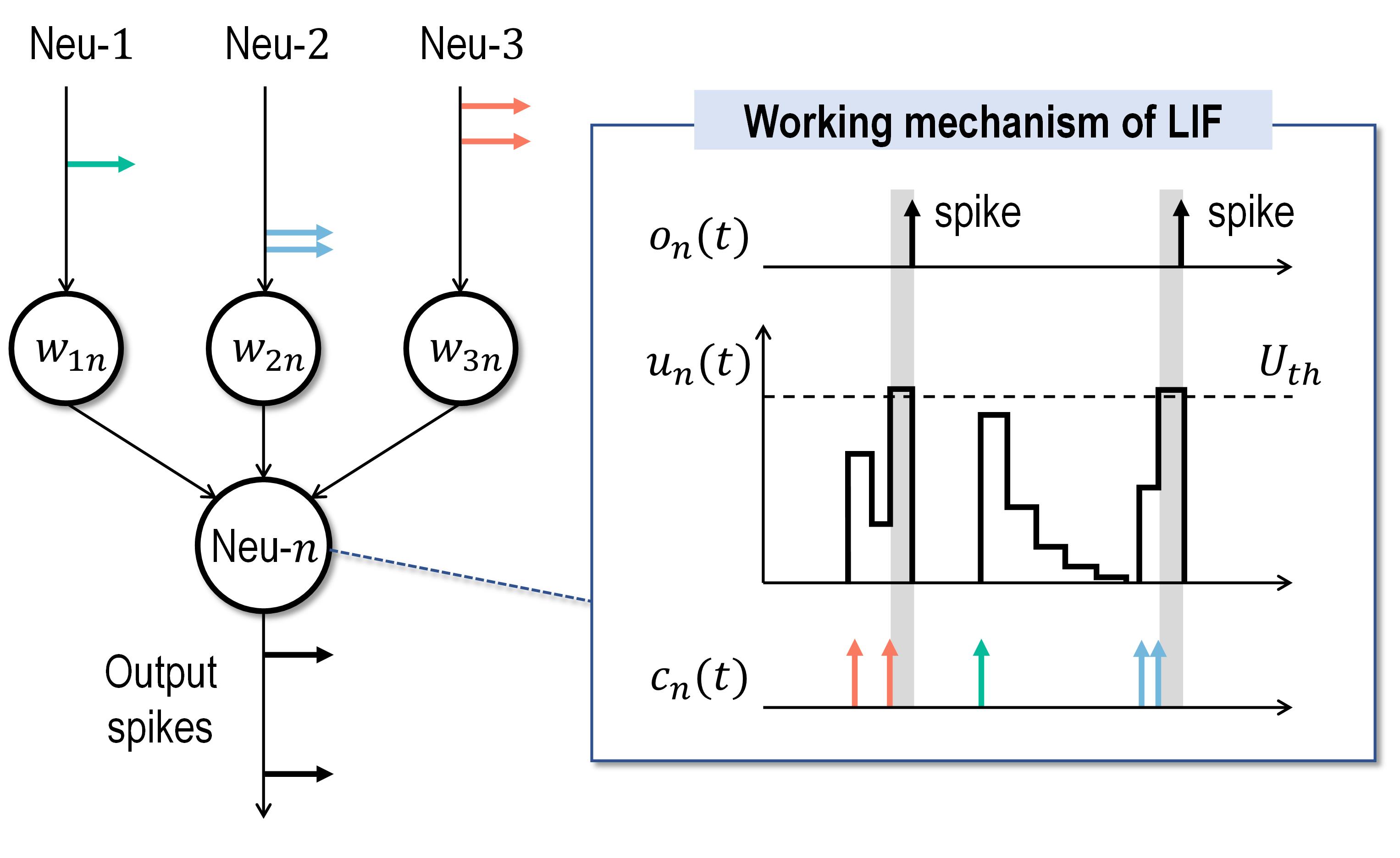}
	\vspace{-2mm}
	\caption{\rmfamily \fontsize{8pt}{0} An illustrative example of the LIF neuron and its working mechanism. The spikes from pre-neurons are first weighted and then fed into the target neuron-$n$, charging the internal membrane potential $u_n(t)$. Spikes will be fired whenever $u_n(t) > U_{th}$. Due to the leakage mechanism, the LIF neuron is able to filter out the isolated spikes, \eg the noise events scattered out in spatio-temporal dimensions.}
	\label{LIFpic}
	\vspace{-2mm}
\end{figure}
\begin{figure*}
	\centering
	\vspace{-2mm}
	\includegraphics[width=0.9\textwidth,height=0.313\textwidth]{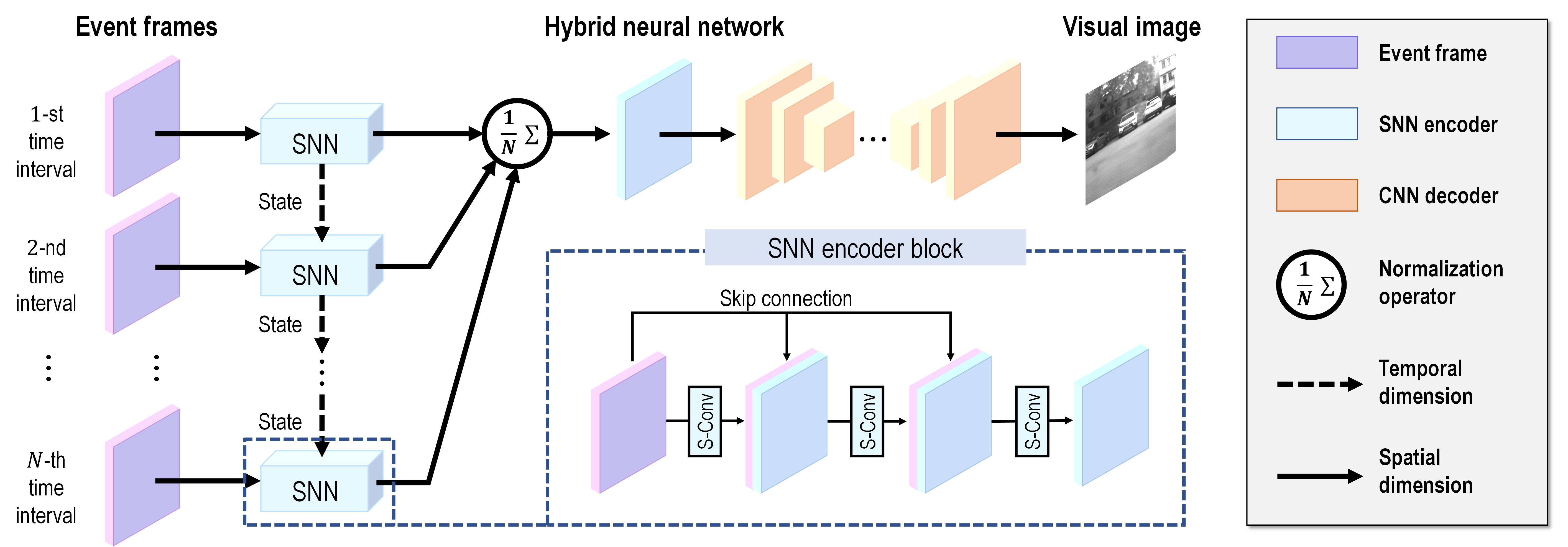}
	\caption{Structure of the hybrid encoder-decoder network. The spatio-temporal information of events is first encoded by SNN blocks, and then transformed to visual images by the CNN decoder.  To reduce the information loss of events, we add skip connections between the event frame and the output of the 1-st, 2-nd spiking convolution (S-Conv) layers.}
	\label{Hyrbrid}
	\vspace{-2mm}
\end{figure*}
\subsection{Reconstruction with a Hybrid Network}\label{sec3-2}
According to Eq.~(\ref{logdiff}), the brightness intensity of the occluded scene is closely related to the number of events. Thus the visual image of the occluded scene can be recovered by event accumulation after the refocusing process without removal of noises. Even though CNN-based methods can be further exploited to alleviate the noise problem, the temporal information behind events cannot be effectively used. In view of this, we propose a hybrid neural network composed by a SNN encoder and a CNN decoder, where both spatial and temporal information of events can be efficiently considered and utilized.
\textbf{SNN Encoder:} Although the noise events are dispersed during the refocusing process, their presence still affects the quality of reconstruction.
To deal with it, we implement the SNN encoder using the leaky integrate-and-fire (LIF) model \cite{wuDirectTrainingSpiking2019}.
As shown in Fig.~\ref{LIFpic}, LIF neurons are usually activated when receiving more continuous spikes. If no new spikes are fed, the internal membrane potential will gradually leak over time. Recall that all signal events are successfully aligned during the refocusing process, \ie they appear more continuously in the temporal dimension, while these noise events are scattered in both time and space. Thus, the leakage mechanism of LIF neuron is able to eliminate the influence of noise events, meanwhile preserving the information of occluded targets.
\par 
\emph{LIF Neuron:} Define $u_n^{l}(t)$ as the membrane potential of  the neuron-$n$ on the $l$-th layer at time $t$. The update of membrane potential can be described as
\begin{equation}\label{tmp-mem-up}
u_n^{l}(t) = \alpha u_n^{l}(t-1) + c_n^{l}(t),
\end{equation}
where $\alpha\in[0,1]$ denotes the decay factor and $c_n^{l}(t)$ is the input current corresponding to neuron-$n$.  Considering the convolution operation in spiking layers, Eq.~(\ref{tmp-mem-up}) can be reformulated as:
\begin{equation}\label{mem-up}
u_n^{l}(t) = \alpha u_n^{l}(t-1) + \sum_{m}w_{mn}o_m^{l-1}(t-1),
\end{equation}
where $o_m^{l-1}(t-1)$ represents the output spike of neuron-$m$ on the $(l-1)$-th layer at time $t-1$, and $w_{mn}$ denotes the synaptic weight between neuron-$m$ and neuron-$n$. Further, we add the reset \& fire mechanism into Eq.~(\ref{mem-up}),
\begin{equation}\label{LIF}
u_n^{l}(t) = \alpha u_n^{l}(t-1)(1-o_n^{l}(t-1)) + \sum_{m}w_{mn}o_m^{l-1}(t-1),
\end{equation}
where the output spike $o_n^{l}(t)$ is defined by
\begin{equation}\label{fire}
o_n^{l}(t) = \left\{
\begin{array}{ll}
1, & \text { if } u_n^{l}(t)>U_{th}, \\
0, & \text{otherwise},
\end{array}\right.
\end{equation}
and $U_{th}$ represents the spiking threshold. Eq.~(\ref{LIF}) indicates that the membrane potential of neuron-$n$ is affected by both its own state and the input spikes. If no new spikes are fed, the membrane potential $u_n^l(t)$ will leak at a certain rate related to the factor $\alpha$. In contrast, if the potential $u_n^l(t)$ is charged up to the spiking threshold $U_{th}$, the potential will be immediately reset to the resting potential $U_{rest}=0$ and simultaneously a spike will be emitted to other neurons. 
\par
\emph{SNN Structure:} As illustrated in Fig.~\ref{Hyrbrid}, our SNN encoder consists of a three-layer structure composed of LIF neurons. To make a balance between computational complexity and information integrity, we present a spatio-temporal representation for events. Given a pre-setting number of event frames, \eg $N$, the refocused event sequence are fairly divided into $N$ time intervals. In each interval, an event frame can be generated by accumulating events over time, and each frame contains two channels (positive and negative events). Thus, every input group includes $N$ event frames and the temporal relationship between event frames is retained. Over time, event frames sequentially pass through the spiking layers, and the membrane potential of spiking neurons updates between time intervals. Since noise events are scattered during refocusing, their influence can be gradually leaked out by the potential update of LIF neurons. Therefore, the noise issue is well alleviated, guaranteeing the reconstruction quality of occluded targets. To avoid the vanishing spike phenomenon in deep spiking layers \cite{pandaScalableEfficientAccurate2020}, we instead implement the decoder with a deep CNN block.

\textbf{CNN Decoder:} Due to the inherent difference between the event feature map and the visual image, we regard the decoding process as a style-transfer task. Here, we adopt the decoder architecture from the generator network used in \cite{zhu2017unpaired}, which shows remarkable results in image style transferring, and adjust the kernel size of the output layer to fit the gray-scale images in our case.
Benefiting from the hybrid structure, the spatio-temporal information of events can be fully utilized by the SNN encoder, and the occluded targets can be effectively reconstructed by the CNN decoder, guaranteeing the overall performance.

\textbf{Training Hybrid Network:}
The synaptic weights in SNN can be trained in a supervised fashion via the spatio-temporal back propagation (STBP) technique \cite{10.3389/fnins.2018.00331,wuDirectTrainingSpiking2019}, where the gradient of each pixel can be derived based on time intervals. And CNNs can be trained via back propagation (BP). Thus the SNN and CNN in the proposed hybrid network can be jointly trained. 

To guide the training, we first exploit the idea of perceptual loss \cite{johnson2016perceptual} for high-level feature learning. 
With a pretrained loss network $\phi$, we denote $\phi_k(X)$ as the output of the $k$-th convolution layer when network $\phi$ processes image $X$. Assume that $\phi_k(X)$ has the shape  $C_{k} \times H_{k}\times W_{k}$, we can formulate the perceptual loss $\mathcal{L}_{per}$ as:
\begin{equation}\label{percep}
\mathcal{L}_{per}(Y,\hat{Y}) = \sum_{k} \frac{\lambda_k}{C_{k}H_{k}W_{k}} \|\phi_k(Y)-\phi_k(\hat{Y})\|_2^{2},
\end{equation}
where $Y$ represents the output of the hybrid network and $\hat{Y}$ is the corresponding ground truth; $\lambda_k$ denotes the weight of the $k$-th feature map. Rather than encouraging the pixel-wise match between images, the perceptual loss encourages the network to learn the similarity between high-level features, leading to better visual results.
\begin{figure*}[t!]
	\centering
	\includegraphics[width=0.95\linewidth]{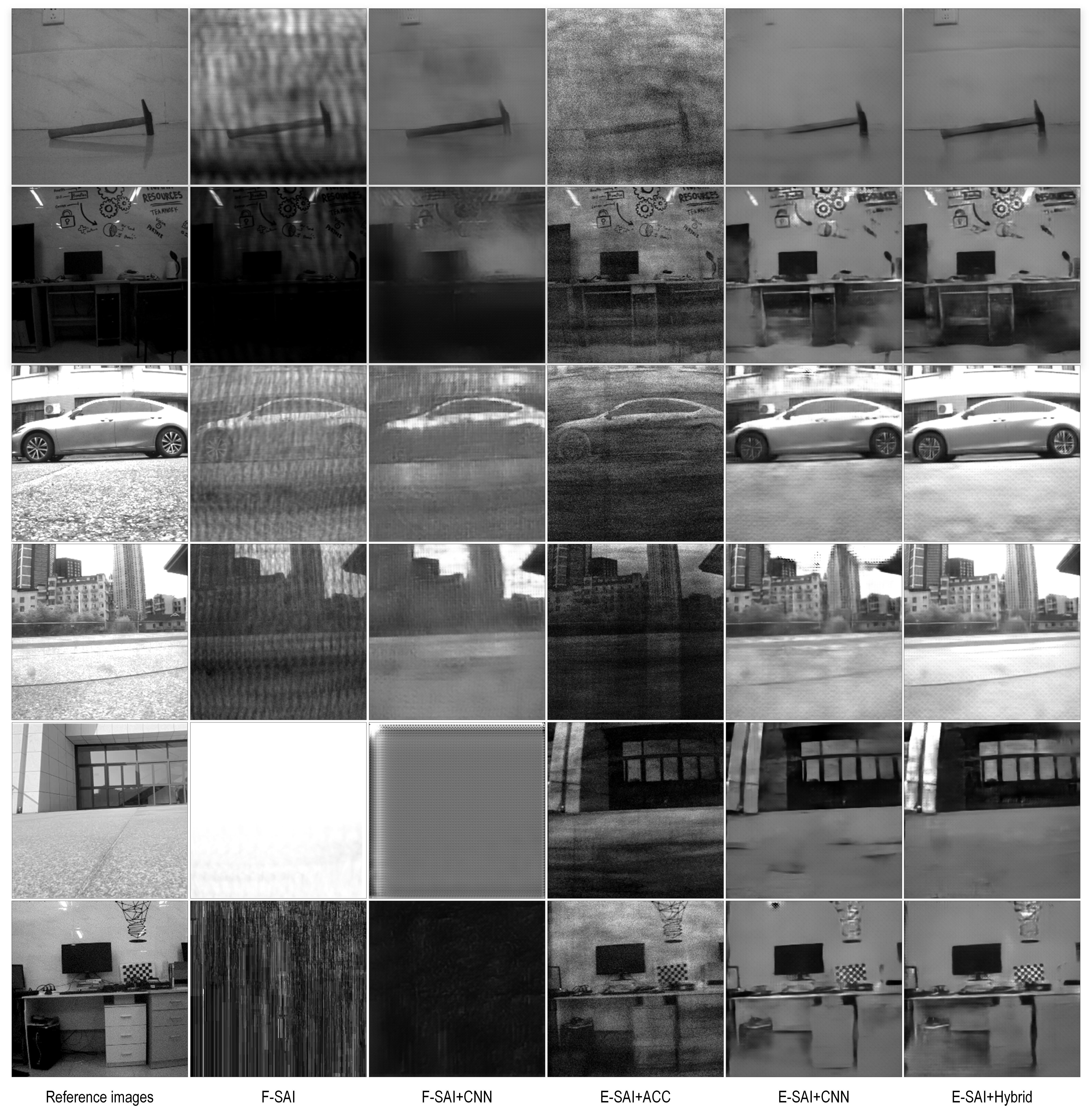}
	\vspace{-3mm}
	\caption{Qualitative comparisons between F-SAI and E-SAI algorithms under very dense occlusions (row 1-4) and extreme lighting conditions (row 5-6) for \emph{indoor} (row 1,2 and 6) and \emph{outdoor} (row 3-5) dataset. } 
	\label{BigVis}
	\vspace{-2mm}
\end{figure*}

\begin{figure*}[t!]
	\centering
	\includegraphics[width=0.99\linewidth]{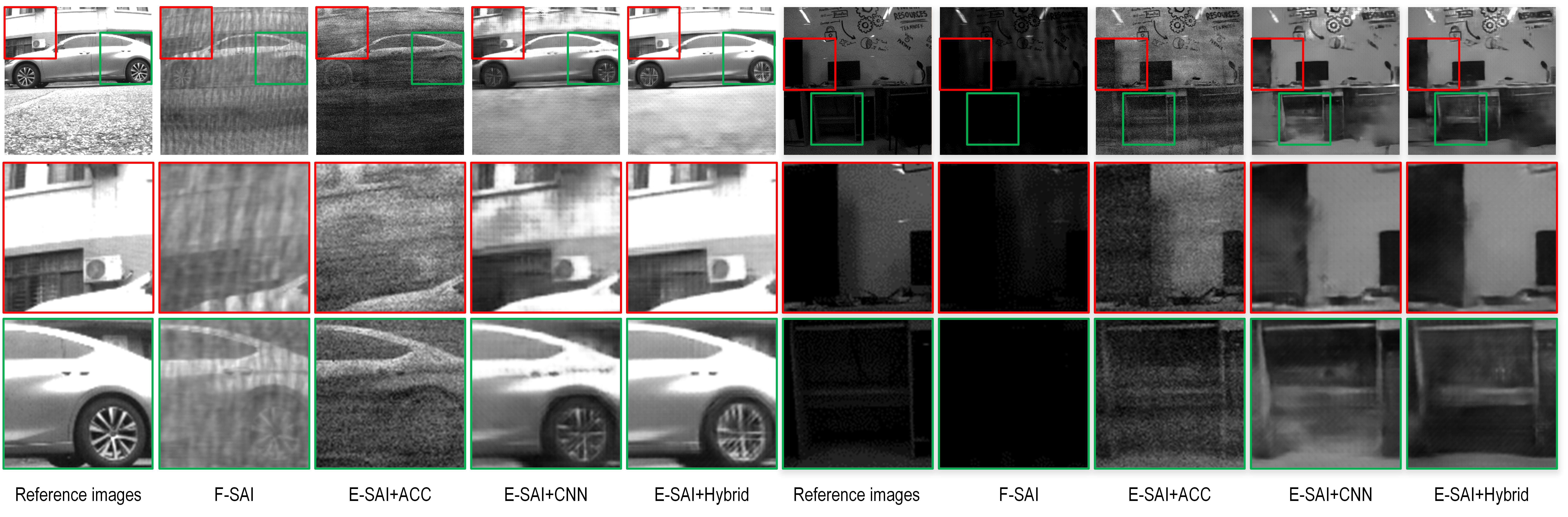}
	\vspace{-2mm}
	\caption{Comparisons of F-SAI and E-SAI with different reconstruction methods. Details are zoomed in for better view.}
	\label{DetailVis}
	\vspace{-1mm}
\end{figure*}

\begin{table*}[t!]
	\centering
	\small
	\begin{tabular}{c|cc|cc|cccc}
		\hline
		\multirow{3}{*}{\textbf{Method}} & \multicolumn{4}{c|}{\textbf{Dense Occlusion}} & \multicolumn{4}{c}{\textbf{Extreme Exposure}} \\ \cline{2-9} 
		& \multicolumn{2}{c|}{\textbf{Indoor}} & \multicolumn{2}{c|}{\textbf{Outdoor}} & \multicolumn{2}{c|}{\textbf{Over}} & \multicolumn{2}{c}{\textbf{Under}} \\ \cline{2-9} 
		& \multicolumn{1}{c|}{PSNR} & SSIM & \multicolumn{1}{c|}{PSNR} & SSIM & \multicolumn{1}{c|}{Entropy} & \multicolumn{1}{c|}{STD} & \multicolumn{1}{c|}{Entropy} & STD \\ \hline
		F-SAI \cite{vaishUsingPlaneParallax2004} & 13.89 & 0.4482 & 10.96 & 0.3124 & 4.855 & \multicolumn{1}{c|}{16.86} & 5.022 & 26.06 \\
		F-SAI+CNN & 26.44 & 0.8077 & 13.93 & 0.3744 & 5.684 & \multicolumn{1}{c|}{20.77} & 4.165 & 4.264 \\ \hline
		E-SAI+ACC & 14.71 & 0.2272 & 8.654 & 0.1887 & 5.706 & \multicolumn{1}{c|}{41.16} & 5.314 & \textbf{50.10} \\
		E-SAI+CNN & 31.27 & 0.8373 & 17.62 & 0.5237 & 6.921 & \multicolumn{1}{c|}{56.16} & 6.105 & 33.84 \\
		E-SAI+Hybrid & \textbf{33.04} & \textbf{0.8429} & \textbf{24.02} & \textbf{0.6524} & \textbf{7.417} & \multicolumn{1}{c|}{\textbf{61.95}} & \textbf{6.204} & 34.87 \\ \hline
	\end{tabular}
	\vspace{-2mm}
	\caption{Quantitative comparisons between F-SAI and E-SAI algorithms. PSNR(dB) and SSIM are exploited for dense occlusion cases. No-reference metrics, \ie 2D entropy and STD are exploited for extreme exposure cases due to the absence of the corresponding APS frames.}
	\label{table}
\end{table*}

\par 
In the pixel level, we add the pixel loss $\mathcal{L}_{pix}$ to maintain the similarity in low-level features like shape and texture. We express the pixel loss as:
\begin{equation}\label{pixel}
\mathcal{L}_{pix}(Y,\hat{Y}) =\frac{\|Y-\hat{Y}\|_1}{CHW},
\end{equation}
where $C \times H\times W$ represents the shape of $Y$ and $\hat{Y}$. Besides, the total variance loss $\mathcal{L}_{tv}(Y)$ in \cite{mahendran2015understanding} is exploited to encourage the spatial smoothness of reconstruction. Thus, the total loss can be summarized as follows.
\begin{equation}\label{total-loss}
\mathcal{L}(Y,\hat{Y}) = \beta_{per} \mathcal{L}_{per}(Y,\hat{Y}) + \beta_{pix} \mathcal{L}_{pix}(Y,\hat{Y}) + \beta_{tv} \mathcal{L}_{tv}(Y),
\end{equation} 
where $\beta_{per}, \beta_{pix}$ and $\beta_{tv}$ are the weights that control the importance of the corresponding loss function.
\section{Experiments and Analysis}\label{chapter-4}
\subsection{Experimental Settings}
\textbf{Event-based SAI Dataset:} We build an event-based SAI dataset where the event streams are captured by a DAVIS346 camera \cite{lichtsteiner128Times1282008} installed on a programmable sliding trail. A large variety of targets are considered including printed pictures,  simple objects and real scenes. They are occluded by the wooden fence installed parallel to the sliding trail to imitate the very dense occlusions, as shown in Fig.~\ref{pipline}. By linearly sliding the DAVIS346 camera, the events triggered by the brightness difference between the wooden fence and the occluded targets are collected. 
The dataset can be divided into two categories according to the shooting scenes: \emph{indoor} and \emph{outdoor}. The \emph{indoor} dataset contains printed pictures, simple objects and real scenes, while the \emph{outdoor} dataset only contains real complex scenes. The gray-scale images are captured simultaneously as collecting the events by DAVIS346 camera since it can output both events and APS (active pixel sensor) frames. Moreover, we collect the APS frames without occlusions and take them as the ground truth of the occluded targets. In summary, the event-based SAI dataset is built with 300 groups of data including 250 groups for \emph{indoor} and 50 groups for \emph{outdoor}, and each group contains a stream of events, a series of APS frames with occlusions and one APS frame without occlusions. For the extreme lighting scenes, there is no APS frame without occlusions due to the over/under exposure problem.

\par 
\textbf{Training Details:} We augment the event-based SAI dataset by flipping (horizontal, vertical, and horizontal-vertical) and rotating (random angles ranging from -10 to 10 degree). Finally, 216 groups (180  \emph{indoor} groups and 36 \emph{outdoor} groups) are augmented to 1296 groups for training, while the rest in dataset are left for the testing phase. 
All networks are trained on NVIDIA TITAN RTX GPUs with batch size 8 for around 500 epochs, and the Adam optimizer \cite{kingma2014adam} is used, where the initial learning rate is set as $5\times {10}^{-4}$ and the step decay learning rate schedule is applied after the 250 epochs.
The 16-layer VGG network \cite{simonyan2014very} pretrained on the ImageNet dataset \cite{russakovsky2015imagenet} is employed as the loss network, where the perceptual loss is calculated at the 2-nd, 4-th, 7-th and 10-th convolution layers.
\par 
\textbf{SAI Methods:} For the frame-based SAI, the approach proposed in \cite{vaishUsingPlaneParallax2004} (\textbf{F-SAI}) is employed, where 35 images are collected with frame-based cameras from different viewpoints. In addition, we design a learning-based F-SAI using CNNs (\textbf{F-SAI+CNN}) with the same 35 images. For the event-based SAI, we evaluate three different reconstruction 
methods, including accumulating method (\textbf{E-SAI+ACC}), pure CNN method (\textbf{E-SAI+CNN}) and the proposed hybrid network (\textbf{E-SAI+Hybrid}). In the E-SAI+CNN method, the refocused event streams are stacked as a $2N$-channel tensor ($2$ represents the polarity) for network input. 
To evaluate the effectiveness of our hybrid network, a pure CNN counterpart is designed by simply replacing the SNN encoder with a 3-layer CNN which has the same network setting as the SNN. By applying the pure CNN model to \textbf{F-SAI+CNN} and \textbf{E-SAI+CNN}, we can fairly compare these learning-based SAI methods.

\subsection{Qualitative Analysis}
As shown in Fig.~\ref{BigVis}, the reconstruction results of F-SAI methods are severely contaminated by dense occlusions where a lot of details are missing. For the extreme lighting scenes, the performance is even worse since the light from the occluded target cannot be correctly measured due to the over/under exposure problems encountered with frame-based cameras. On the contrary, E-SAI methods are able to produce results with better visual effects and retain more details. Due to the inherent high dynamic range of event camera, E-SAI methods do not suffer from over/under exposure problems, thus the occluded target can be effectively reconstructed under extreme lighting conditions.
\par
To reveal the advantages of our hybrid network, we compare the visual results of E-SAI with different reconstruction techniques in Fig.~\ref{DetailVis}. It is obvious that the result of E-SAI+ACC is often noisy because both signal and noise events are indiscriminately accumulated for reconstruction. In the learning-based approaches, the E-SAI+CNN fed directly with the stacked event frames cannot efficiently deal with the temporal information of asynchronous events, and thus degrading the visual quality with detail losses, artifacts and saturation. But these issues can be mitigated by the hybrid architecture, where the temporal information is utilized by the SNN encoder. Over time, LIF spiking neurons can efficiently leak out the influence of noise events which are either emitted randomly 
or scattered out after the refocusing process. Consequently, the proposed hybrid network generates images with the best visual quality.

\subsection{Quantitative Analysis}
In Table \ref{table}, we evaluate the quantitative results of the proposed system. In the dense occlusion experiment, the metrics PSNR and SSIM \cite{wang2003multiscale} are employed for quantitative comparison, where the aligned APS images captured by DAVIS346 are considered as the ground truth. 
In the extreme exposure part, the no-reference assessment metrics two-dimensional (2D) entropy \cite{zhu2020retina} and standard deviation (STD) are exploited to evaluate the image quality. 2D entropy measures the amount of image information and higher value indicates more information. STD is used to assess the contrast of image and larger value means higher contrast. 
\par 
Exploiting learning-based techniques, F-SAI+CNN is able to produce better results than F-SAI under dense occlusions. But both frame-based methods cannot deal with the over/under exposure problem due to the low dynamic range of traditional camera. On the contrary, event-based approaches can effectively reconstruct visual images with more information and better contrast. However, it is hard for E-SAI+ACC to produce satisfactory PSNR and SSIM results since the emission of events is based on the brightness change in the logarithmic domain, which differs from the intensity directly recorded in reference images. Through learning the mapping relationship between the event domain and the image domain, this problem can be well solved by the E-SAI+CNN and E-SAI+Hybrid. 
Regarding the learning-based E-SAI, the hybrid network excels its pure CNN counterpart over 6 dB in PSNR and 0.12 in SSIM under complex outdoor scenes. This demonstrates that the use of SNN encoder not only achieves the denoising purpose, but also maintains the integrity of overall structure. In summary, our E-SAI+Hybrid method largely outperforms other algorithms under dense occlusions, and can produce more natural visual results in extreme lighting environments.




\section{Conclusion}
In this work, we proposed a novel SAI algorithm based on event cameras. With extremely low latency and high dynamic range of event cameras, our method is able to handle the disturbance of dense occlusion and does not suffer from the over/under exposure problem. This greatly expands the usage of SAI algorithm, enabling the application under harsh conditions like daytime astronomical observation and nighttime penetrating imaging. Moreover, a hybrid SNN-CNN network is proposed to process the output of event camera. Benefiting from the combination of SNN and CNN, the spatio-temporal information of events is well utilized and the reconstruction quality of occluded targets is guaranteed.
To test our method, we build an event-based SAI dataset including scenes under heavy occlusions and extreme lighting conditions. The result verifies that our approach is effective to these harsh environments and can reconstruct the occluded target with impressive visual effects.




\begin{table*}[h]
	\centering
	{\Large \bf 
		{Event-based Synthetic Aperture Imaging with a Hybrid Network
			\\
			- Supplementary Material -}
			\par}
		\vskip .5em
		\vspace*{12pt}
\end{table*}

\setcounter{section}{0} 
\setcounter{figure}{0} 
\setcounter{table}{0} 
\begin{figure}[htb]
	\centering
	\includegraphics[width=\linewidth]{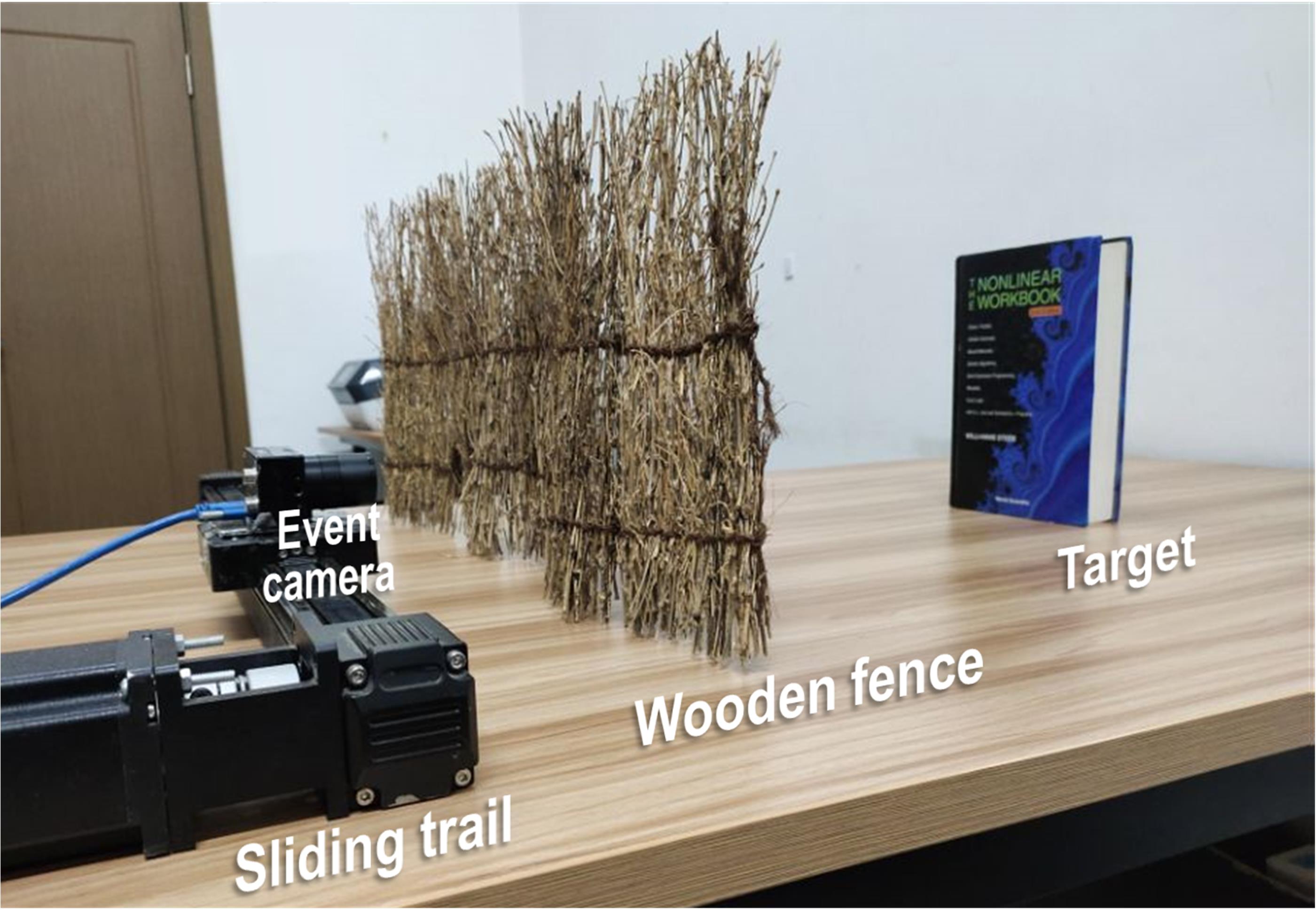}
	\caption{An example of our experimental scenes. It consists of occlusions (wooden fence), targets (book) and an event camera installed on a programmable sliding trail.} 
	\label{scene}
\end{figure}
\section{Experimental Scenes}
As displayed in Fig.~\ref{scene}, we install the event camera on a programmable sliding trail and employ a wooden fence to simulate the densely occluded scenes. When the camera moves linearly on the sliding trail, the events triggered by the brightness difference between occlusions and targets can be collected from different viewpoints.

\begin{figure}[htb]
	\centering
	\begin{subfigure}{0.45\textwidth}
		\includegraphics[width=\textwidth]{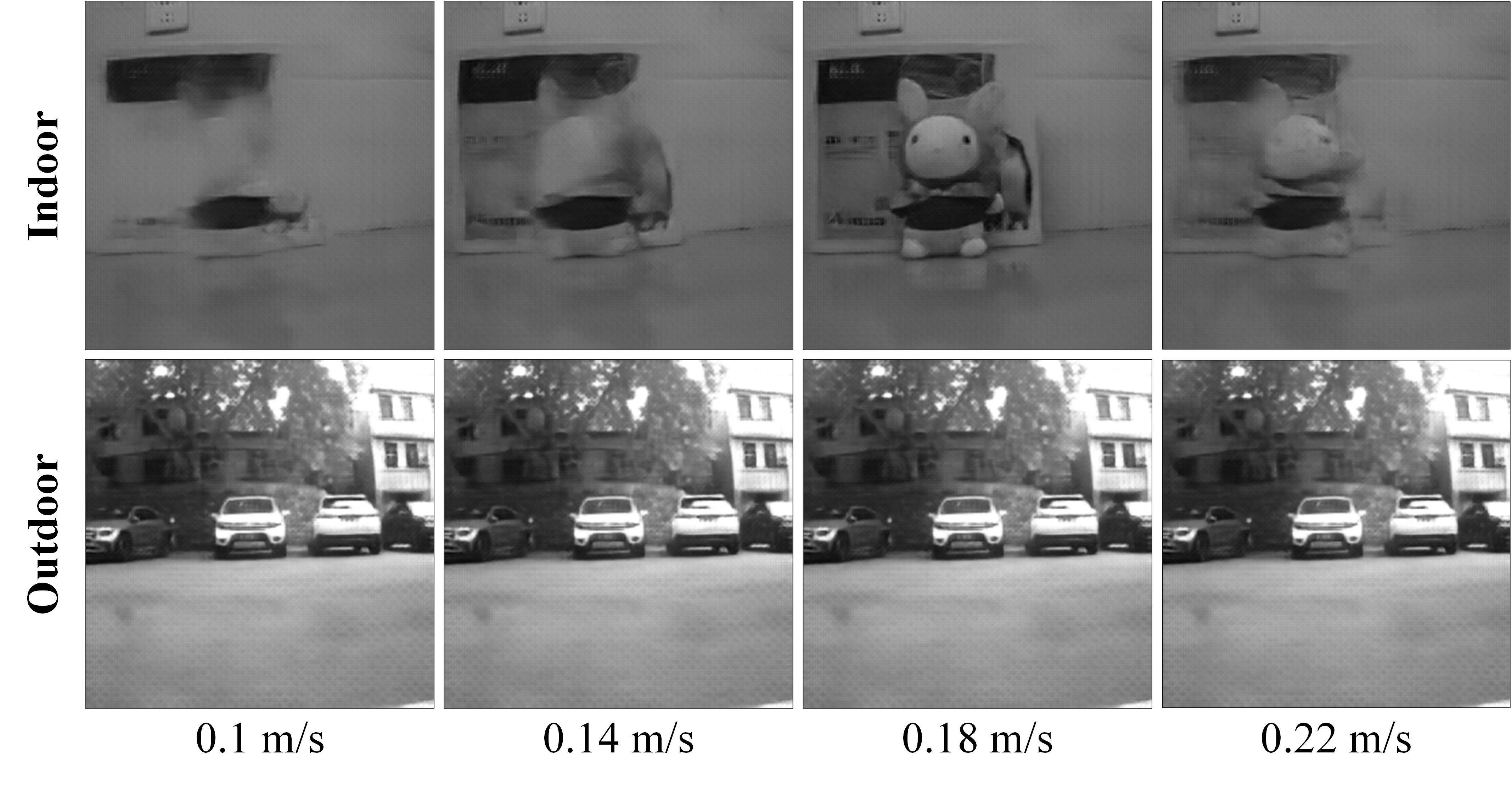}
		\caption{\rmfamily \fontsize{8pt}{0} Reconstruction results}
		\label{speedres}
	\end{subfigure}
	\begin{subfigure}{0.23\textwidth}
		\includegraphics[width=\textwidth]{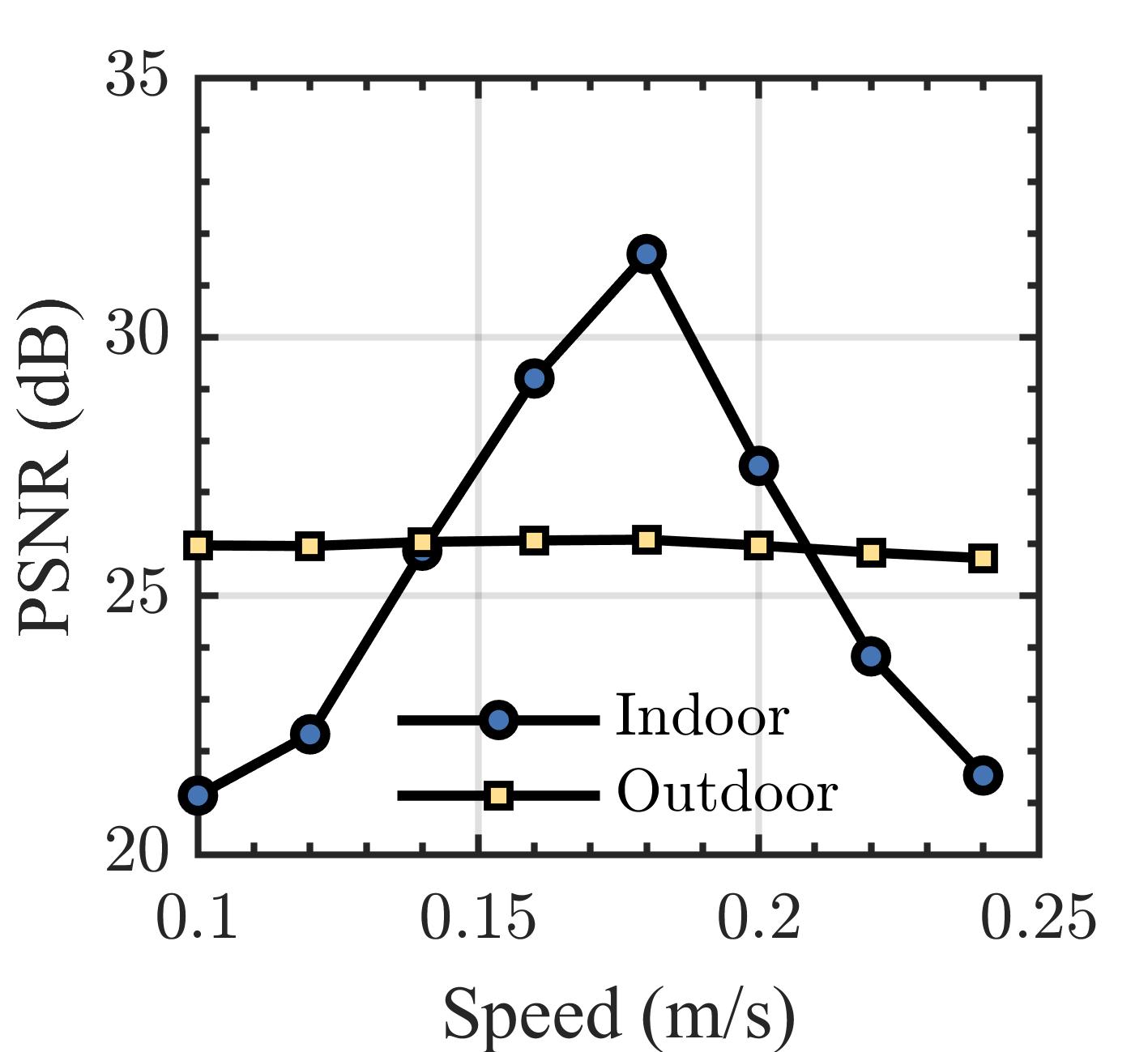}
		\caption{\rmfamily \fontsize{8pt}{0} PSNR results}
		\label{speed-psnr}
	\end{subfigure}
	\begin{subfigure}{0.24\textwidth}
		\includegraphics[width=\textwidth]{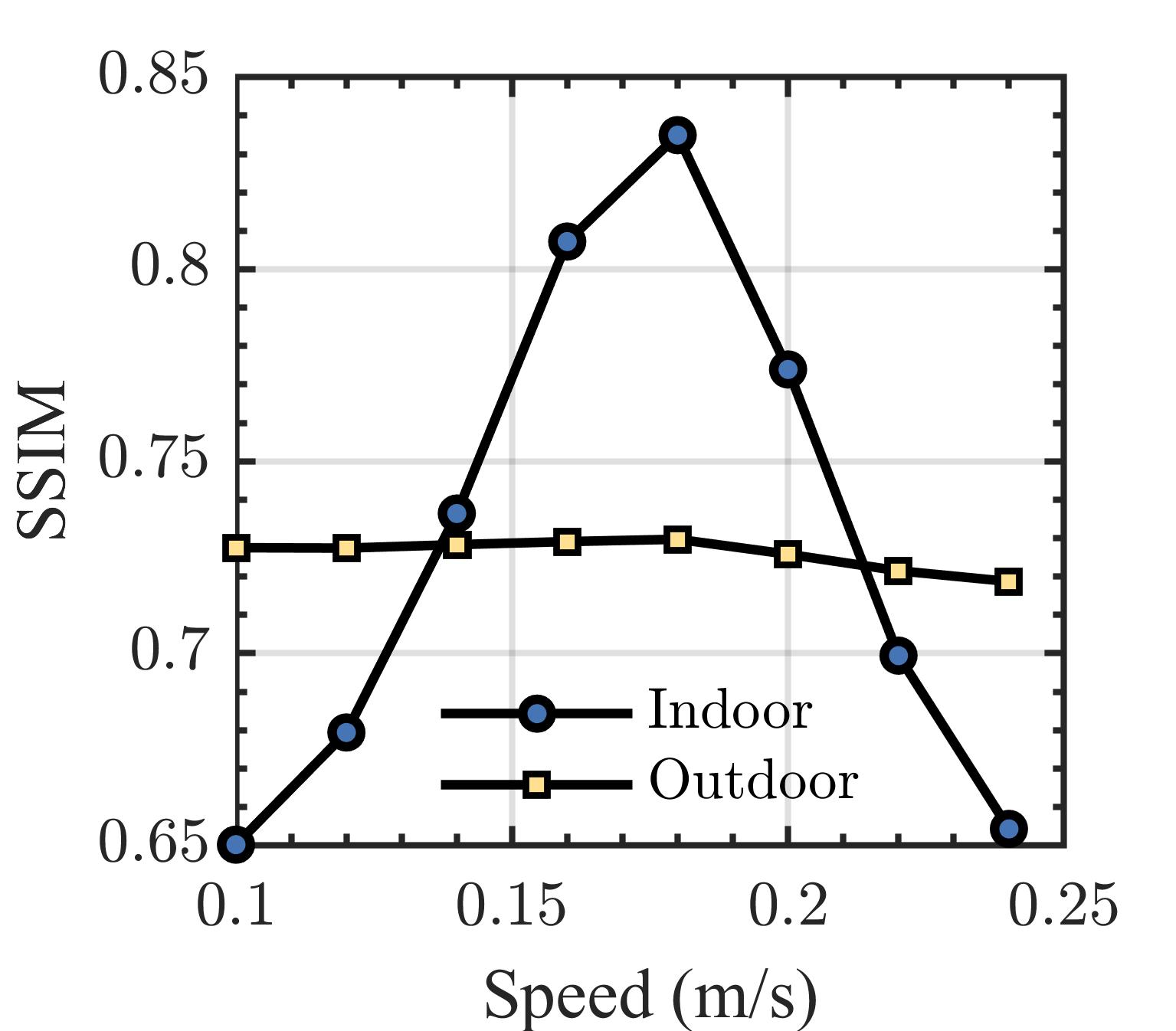}
		\caption{\rmfamily \fontsize{8pt}{0} SSIM results}
		\label{speed-ssim}
	\end{subfigure}
	\caption{Influence of camera motion accuracy. (a) Reconstruction results of indoor and outdoor data under different estimates of camera speed. (b), (c) Quantitative results. The real camera speed is $0.177\ m/s$.} 
	\label{speed}
\end{figure}
\begin{figure*}[h]
	\centering
	\includegraphics[width=0.97\linewidth]{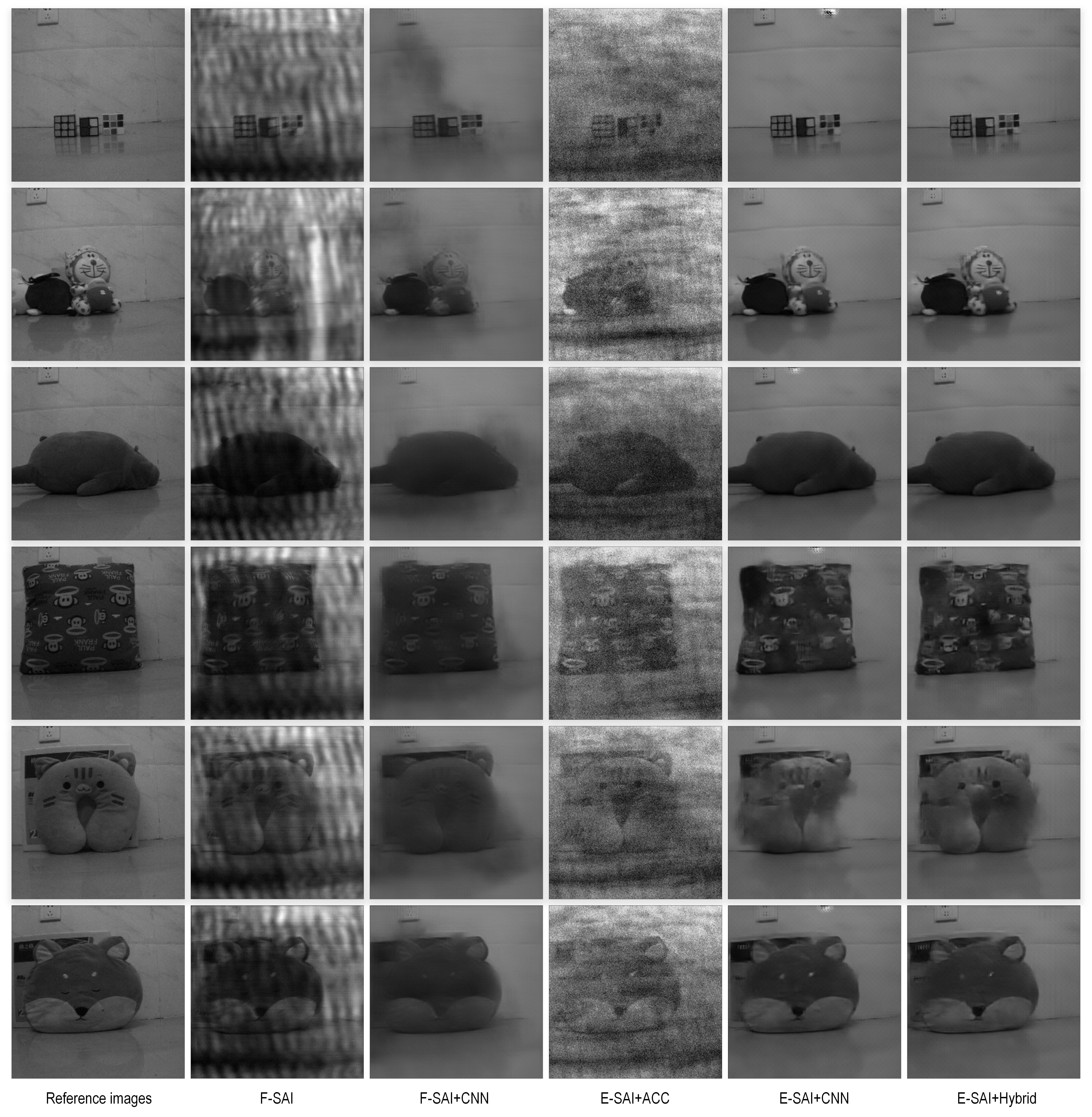}
	\caption{Qualitative comparisons between F-SAI and E-SAI algorithms on the \emph{indoor} dataset. } 
	\label{Addindoor}
\end{figure*}
\begin{figure*}
	\centering
	\includegraphics[width=0.97\linewidth]{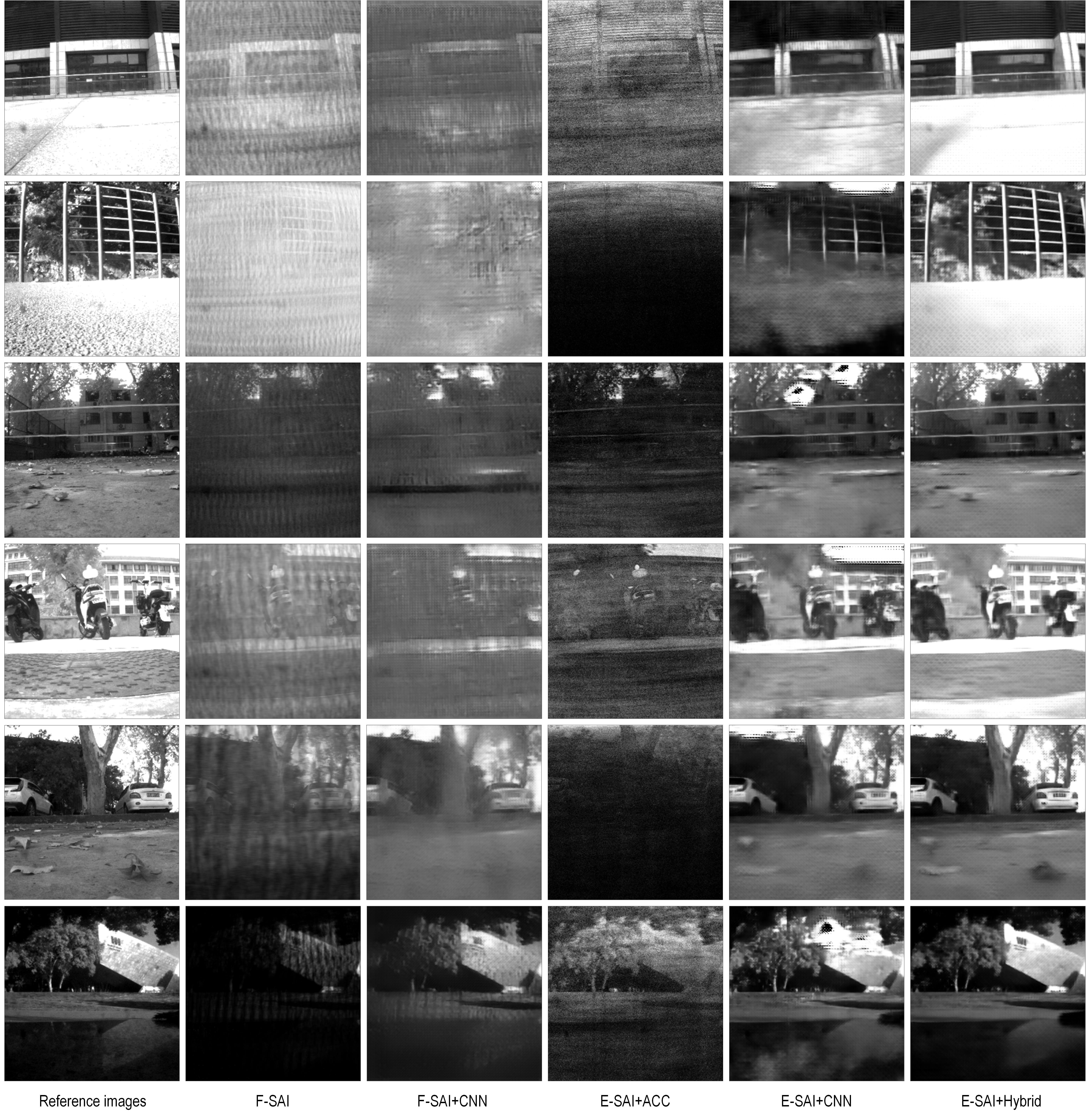}
	\caption{Qualitative comparisons between F-SAI and E-SAI algorithms on the \emph{outdoor} dataset.} 
	\label{Addoutdoor}
\end{figure*}

\section{Extra Experimental Results}
\subsection{Influence of Camera Motion Accuracy}
Our experiments are mainly based on one-dimensional uniform camera motion, thus the estimation of the camera speed is directly related to the refocusing module and the overall performance of our proposed method. To investigate this, we choose two pairs of data (from indoor and outdoor datasets each) and apply E-SAI+Hybrid to them with camera speeds varying from $0.1\ m/s$ to $0.24\ m/s$. Note that the actual camera speed is $0.177\ m/s$.
\par
As illustrated in Fig.~\ref{speed}, the reconstruction quality of indoor data is severely degraded when the speed estimation error is large. Since the targets in our indoor dataset are often closer to the camera (\ie close-view targets), estimation error will cause a significant shift of targets on the imaging plane. Thus these signal events cannot be reliably aligned during refocusing and may be treated as noise during reconstruction, leading to serious blur and missing details in final results. On the contrary, our outdoor dataset mainly contains far-view targets, and thus is less sensitive to the estimation error.

\subsection{Analysis of Indoor and Outdoor Results}
In the experiments on indoor dataset, we mainly test F-SAI and E-SAI methods with simple objects. As displayed in Fig.~\ref{Addindoor}, the results of F-SAI and E-SAI+ACC are often blurry and noisy since the light information of both occlusions and targets are equally treated during reconstruction. For the learning-based SAI, F-SAI+CNN is able to recover the shape of targets and achieve a better de-occlusion effect than F-SAI, but the result still suffers from the issues of detail losses and artifacts. On the other hand, it is hard for E-SAI+CNN to simultaneously handle spatial and temporal information inside events, thus the reconstruction quality is often degraded by the disturbance of noise events. 
\par 
For the outdoor dataset, we consider more general targets including cars, fields and buildings. Compared with the indoor scenes, outdoor lighting conditions are much more complicated, making it harder for frame-based SAI methods (F-SAI and F-SAI+CNN) to generate clean results, as shown in Fig.~\ref{Addoutdoor}.
Similarly, complex lighting conditions also degrade the performance of E-SAI due to the increase of noise events, \eg the events triggered by the brightness change of occlusions $\mathcal{E}^{OO}_\theta$ and occluded scenes $\mathcal{E}^{AA}_\theta$. The rising number of noise events not only makes the target indistinguishable in the results of E-SAI+ACC, but also brings more disturbances to E-SAI+CNN, deteriorating the reconstruction quality with serious saturation problem. Thanks to the hybrid SNN-CNN architecture, the issue of noise events can be alleviated from the temporal dimension. Therefore, our E-SAI+Hybrid is more robust to complex lighting conditions compared to other SAI methods, and can achieve the best visual effects on both indoor and outdoor datasets.

\section{Extra Information}
\subsection{Implementation Details}
Each data sequence lasts about 0.7 seconds in our datasets. In our experiments, we divide each sequence into 30 time intervals, \ie $N=30$, for 
E-SAI+CNN and E-SAI+Hybrid to make the input information equal. In network training, we set the loss weights as $\left[\beta_{per},\beta_{pix},\beta_{tv}\right]=\left[1,32,2e{-4}\right]$. For the perceptual loss, we set the weights $\left[\lambda_2,\lambda_4,\lambda_7,\lambda_{10}\right]=\left[1e{-1},1/21,10/21,10/21\right]$.

\subsection{Network Architectures}
Let $\rm cCsS$-$\rm K$ denotes a $\rm C \times C$ Convolution-BatchNorm-ReLU layer with stride $\rm S$ and $\rm K$ kernels. $\rm r$-$\rm K$ denotes a residual block composed by a $\rm c3s1$-$\rm K$ layer and a $3\times 3$ Convolution-BatchNorm layer with $\rm K$ kernels and stride $1$.
$\rm convCpP$-$\rm K$ denotes a $\rm C \times C$ Convolution layer with stride $1$, padding $\rm P$ and $\rm K$ kernels. $\rm deconv$-$\rm K$ denotes a $3\times 3$ fractional-strided-Convolution-BatchNorm-ReLU layer with $\rm K$ kernels and stride $1/2$.
\par 
Then, the CNN decoder consists of: $\rm c7s1$-$64$, $\rm c3s2$-$128$, $\rm c3s2$-$256$, $\rm r$-$256$, $\rm r$-$256$, $\rm r$-$256$, $\rm r$-$256$, $\rm r$-$256$, $\rm r$-$256$, $\rm r$-$256$, $\rm r$-$256$, $\rm r$-$256$, $\rm deconv$-$128$, $\rm deconv$-$64$,  $\rm c7s1$-$1$. Note that in the output layer $\rm c7s1$-$1$, we replace ReLU function with Tanh function to normalize the output and do not use batch normalization. With the same CNN decoder, these learning-based SAI methods in our experiments can be described as:
\begin{itemize}
	\item \textbf{F-SAI+CNN:}  $\rm conv3p1$-$\rm 16$,  $\rm conv1p0$-$\rm 16$,  $\rm conv1p0$-$\rm 32$ followed by the CNN decoder.
	\item \textbf{E-SAI+CNN:}  $\rm conv3p1$-$\rm 16$,  $\rm conv1p0$-$\rm 16$,  $\rm conv1p0$-$\rm 32$ followed by the CNN decoder.
	\item \textbf{E-SAI+Hybrid:}  $\rm Sconv3p1$-$\rm 16$,  $\rm Sconv1p0$-$\rm 16$,  $\rm Sconv1p0$-$\rm 32$ followed by the CNN decoder, where $\rm SconvCpP$-$\rm K$ denotes a $\rm C \times C$ spiking-Convolution layer with stride $1$, padding $\rm P$ and $\rm K$ kernels.
\end{itemize}
For fair comparison, we also add skip connections between the input tensor and the output of the 1-st, 2-nd Convolution layers in \textbf{F-SAI+CNN} and \textbf{E-SAI+CNN}. 

\newpage
{\small
	\bibliographystyle{ieee_fullname}
	\bibliography{library}
}

\end{document}